# Personalized fall detection monitoring system based on learning from the user movements

Pranesh Vallabh[1], Nazanin Malekian[2,1], Reza Malekian[1], Ting-Mei Li[3]
[1]Department of Electrical, Electronic and Computer Engineering, University of Pretoria
Pretoria, South Africa
[2] Department of Social Communication, East Tehran branch, Islamic Azad university, Tehran, Iran
[3]Department of Electrical Engineering, National Dong Hwa University, Hualien, Taiwan

**Abstract**

Personalized fall detection system is shown to provide added and more benefits compare to the current fall detection system. The personalized model can also be applied to anything where one class of data is hard to gather. The results show that adapting to the user needs, improve the overall accuracy of the system. Future work includes detection of the smartphone on the user so that the user can place the system anywhere on the body and make sure it detects. Even though the accuracy is not 100% the proof of concept of personalization can be used to achieve greater accuracy. The concept of personalization used in this paper can also be extended to other research in the medical field or where data is hard to come by for a particular class. More research into the feature extraction and feature selection module should be investigated. For the feature selection module, more research into selecting features based on one class data.

## 1 Introduction

One-third of the elderly population aged 65 years or more fall at least once each year, whereas half of the elderly population older than 80 years fall each year [1], [2], [3]. The increase in elderly population, notably in developed countries, and the number of elderly people living alone can result in increased healthcare costs which can cause a huge burden on the society and individuals [2], [4], [5], [6]. Due to the shortage of nursing homes, more elderly people are required to stay at home [7]. Fall detection systems are used by elderly people who live alone and cannot alert anyone for help when a fall occurs if they sustain serious injuries or if they become unconscious [8]. Fall detection systems classifiers two sets of data, activities of daily living (ADLs) and fall activities. ADLs are a "wide set of actions characterizing the habits of people, especially in their living places e.g. walking, sitting, standing, and etc." [9]. Fall detection systems have evolved over the past few years, from a button pendant to the following fall detection systems - wearable sensors, ambient sensors, and camera-based sensors. With wearable sensors being the most popular, due to the fact it provides both outdoor and indoor monitoring. However, many fall detection systems make use of experimental/ laboratory data when designing these systems, which results in low accuracy when tested outside the experimental environment. In most studies the systems are trained to detect a small subset of activities. In this paper, personalization techniques in fall detection system are explored which increases the accuracy of the overall system by learning from the user movements better and allows inclusion of new activities. Current research in fall detection systems relies solely on experimental data which make use of young people to perform these activities. The problem is that young people move differently compare to elderly people, which can cause the fall detection system to produce false alarms; since the system does not know the movements of the elderly people.

Duration of falls for elderly people may be longer compared to that of young adults [10]. We are exploring personalization techniques in fall detection to increase the performance, accuracy, and to increase the elderly's independence by not forcing them to perform limited activities. The personalization model will adapt to the person's movements and the lifestyle of the user in terms of the activities the user performs. Thus, personalization model will reduce false alarms and ensure that a fall is detected. The contribution in this paper is the approach for the personalized fall detection system, which provides the following advantages, the inclusion of new activities, eduction in false alarms; and the system does not require fall data.

The remainder of this paper is organized as follows. The common approach is designing of a fall detection system and the problems of the fall detection system approach are presented in section II and III, respectively. Then, in section IV, the personalization fall detection system approach is presented. Moreover, the personalization model and how personalization works are described in section V and VI. Finally the simulation model and simulation results, experiments and the experimental results are shown.

## 2 Common approach when designing a fall detection system

Corresponding author: reza.malekian@ieee.org

Figure 1 shows the most common fall detection approach which is used in many studies when designing a fall detection system. The model comprises of the following parts which will be discussed below: data collection, feature extraction, feature selection, and classifier.

### A. Data collection

The first step is to collect data from sensors. These can be either wearable sensors or ambience sensors and/or camera based sensor systems. The popular one being wearable sensors which have the following advantages: cheap, indoor and outdoor monitoring, and it provides privacy. An example of a wearable sensor is an accelerometer sensor, where the x, y, and z-axis of the accelerometer data is used.

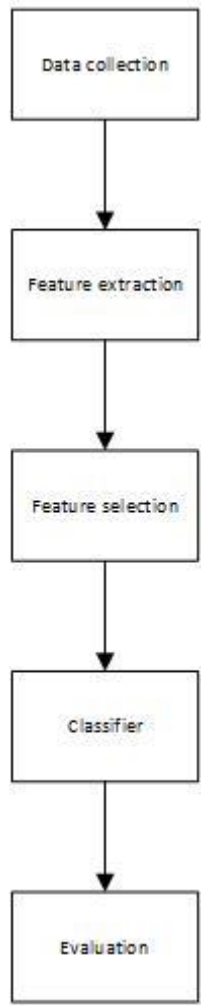

Fig. 1: Fall detection approach.

### B. Feature Extraction

Feature extraction is a method whereby significant attributes are found from raw data which consists of meaningless information which plays a vital part in determining the accuracy of the fall detection system [11], [12],[13], [14], [15]. Fall detection systems require distinctive features to represent the different activities and need to be able to classify falls from ADLs [10]. There are different features, each having relevant characteristics to specific ADLs or fall activity being performed [12]. Features should be carefully selected in order to produce a small descriptive dataset [11]. The dataset's descriptive power is impacted by the number of features that the dataset is comprised of [11]. The disadvantage of feature extraction in fall detection systems, it can increase computational complexity; since it takes time to extract features. Example of feature extraction on the accelerometer data includes the mean of the x-axis, standard deviation on the z-axis, etc.

### C. Feature selection

The more features a database has, the more descriptive it becomes and it becomes difficult to find meaningful relationships among the class as feature space grows exponentially.

The performance of the machine learning algorithm is also dependent on the feature space [11], [12], [15], [16]. By finding features which describe the data better and discarding the redundant features, computational speed and prediction accuracy can be improved [11], [16]. The problem of selecting features from a N dimensional feature space is known as feature selection method [17]. The feature selection algorithms are used to detect and discard features that provide a minimum contribution to the performance of the classifier [12]. Feature selection provides the following advantages; it reduces the cost of pattern detection and the dataset and provides better performance [16], [17]. Feature selection is done once on all the features extracted from the training data; during the testing or the final system, the features from the feature selection method are extracted from the data collected. The biggest problem of feature selection is that when a new data or activities are added to the system, the feature selection method should be run again which increases the computational complexity. This would not work on a smartphone device since there are not enough resources and it does not have a big processing power compared to a computer. One solution is to send the new data to a server, and have the server to send back the new features to be extracted. The feature selection and feature reduction method would not work since each person does not perform the same activity.

### D. Classifiers

The following are used to detect falls, threshold or rule-base algorithms and machine learning algorithms. The basic and earliest fall detection algorithm makes use of the threshold method, where features are extracted from a sensor and compared to a pre-defined value. The advantage of the threshold method is that it is easy to implement, has a low power budget and uses less computational power [9], [18], [19]. The threshold algorithm makes use of the following properties of fall when designing the algorithm: impact of the fall, angle or orientation of the user, and how long the user is in a stationary mode. There are two types of machine learning algorithms: supervised and unsupervised algorithm. Supervised machine learning algorithms make use of label data for training the system and the output of the system is controlled [11], [12]. Popular supervised machine learning algorithms include Naïve Bayes, k-Nearest neighbour, support vector machine, hidden Markov model, and artificial neural network. Unsupervised learning algorithms make use of unlabeled data for training the system [11]. This type of learning algorithm can only be trained on fall data or non-fall data [20]. The classifier can be trained with new activities on the fly. Popular unsupervised classifiers include one-class support vector machine, neares tneighbour, and Gaussian Mixture Model.

**E. Common approach implemented in fall detection system**

Based on the approach, the following model (Model 1) shown in Figure 2 is used in almost all fall detection systems when detecting a fall. The first step is to capture data from fall detection sensors. Relevant features can be extracted from the captured data, to create an input vector. The input vector is inserted into a classifier, where the classifier classifies the input vector as ADL or fall activity. If it is a fall activity, a fall alarm is sounded. The fall alarm will alert the user emergency contacts that a fall has occurred. The process is repeated again.

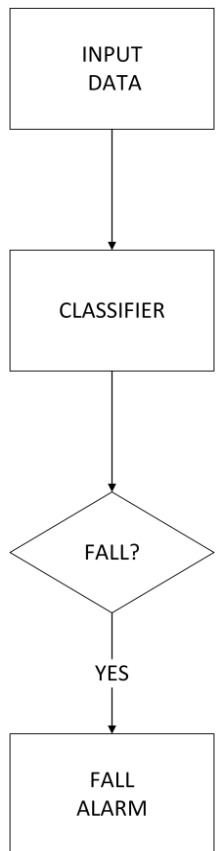

Fig. 2: Model 1 represents the common fall detection model used in most studies.

## 3 Problems of the fall detection system approach

Threshold-based methods include limited recognition ability, low precision, and the pre-defined value cannot address inter-individual difference [8], [19], [21], [22]. The pre-defined value is determined without being based on any theories and/or experiments; which makes it difficult to set a value that can distinguish between ADL and fall activities, also it is difficult to adapt the thresholds to new types of activities [18], [20]. To solve the problems of threshold methods, supervised machine learning algorithms were incorporated - these obtained higher accuracies when compared to the threshold method during laboratory experiments. Certain supervised machine learning algorithms can perform better on certain activities compared to other activities [12]. The problem with supervised methods is that they train with experimental fall activities which do not represent a true fall event and machine learning algorithms have complex implementations rate [19]. In real scenarios, the performance of these fall detections were low, due to the supervised machine learning algorithms making use of simulated fall data which is performed on a soft mattress which does not represent a real fall event (which is usually spontaneous). Since it is difficult to obtain real fall data, about 94% of fall detection studies make use of simulation data [23]. Using artificial fall as training data can result in over-fitting, which causes poor decisions [20]. The simulated falls vary a lot in terms of speed and the nature of fall compare to the real-life falls, and the impact of the fall during experiments is reduced due to the mattress [24], [25]. Both threshold and supervised classifiers cannot provide a user specific solution for each individual user; since each individual has different characteristics and motion patterns compare to the participants used in the training data [21]. Since supervised machine learning algorithms require everyday activities and fall activities to classify, as well as limited fall data (which creates an imbalance i.t.o classification), it is hard for these algorithms to classify accurately. The current research in fall detection is limited to a few activities which restrict a user's movements to certain activities. Fall detection can recognize simple activities such as walking, running, standing and sitting but does not address extreme activities such as housework and outdoor activities. False alarms are not addressed by the fall detection system which can lead to frustration. This frustration can occur when the normal activity being performed by the user is treated as a fall event because that normal activity is not learned by the system.

## 4 PERSONALIZATION FALL DETECTION SYSTEM APPROACH

The c is based on four assumptions:
- Basic smartphone. Most elderly people own a low-cost smartphone. All low-cost smartphone consists of an accelerometer sensor.
- Different ADLs. People perform different types of activities compare to one other. Current fall detection systems are limited to a small set of activities.
- Movement. People movement is different from one other. Elderly people movements are slower compared to young adults.

- Use of only ADL data for training. Current fall detection studies make use of laboratory fall data since real fall data is hard to capture. The problem of fall data, it does not represent a true fall event since the fall is not spontaneous and the participant performing the fall is conscious.

## 5 The personalization model

Figure 3 shows the personalization model design. Model 1 typically uses other people's data except the user's data to train the classifier. The personalization model in Figure 3 make use of the user data to customize the system. The personalized model makes use of unsupervised classifier since it is easier to adapt and re-train. Unsupervised classifier does not need fall data since it would be impossible to obtain from the person unless simulated fall activities are used from other people from the training dataset, which will reduce the quality of the classification system.

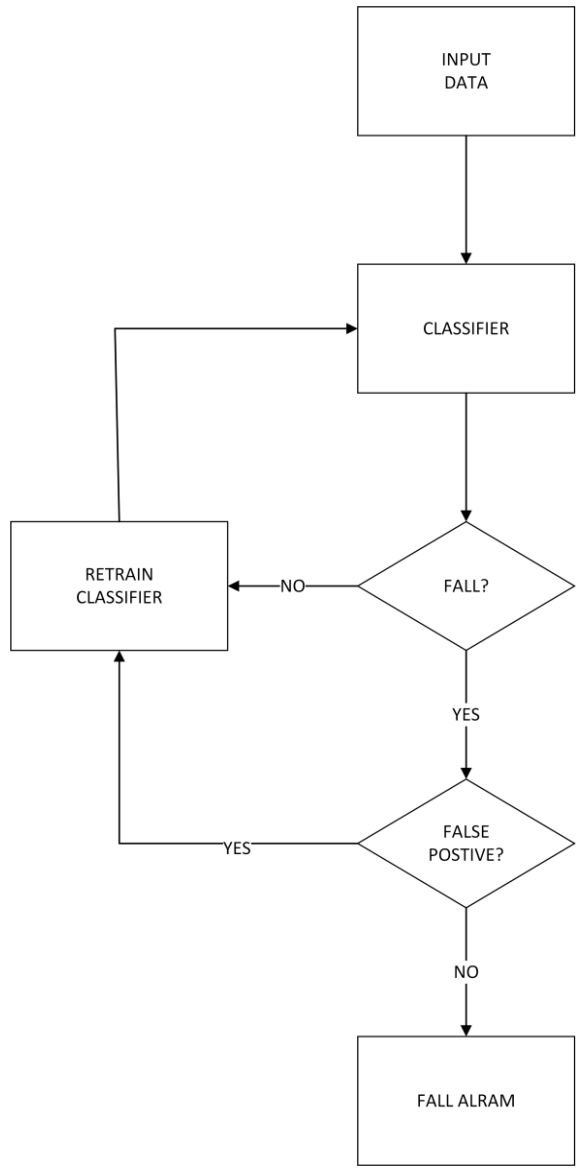

Fig. 3: Personalization model for fall detection system

## 6 How personalization works

If the unsupervised classifier predicts the input data as ADL, the classifier is retrained with the input data. By doing this, the fall detection system will learn to predict the user movements better. Most studies trained the classifier with limited ADL and if the user performed a certain daily activity which is not learnt by the classifier, the classifier will detect it as a fall activity. This will result in high false alarms and irritation among the users. If the unsupervised classifier predicts the input data as a fall, the user is alerted a fall has occurred and has two options on the screen "false alarm" or "fall". If the "false alarm" button is pressed, the classifier is retrained with the input data since an ADL has occurred instead of a fall activity. This will help to reduce the number of false alarmsand learn a new ADL.

## 7 Simulation

A simulation is done to compare the models in Figure 2 and Figure 3. The "tFall" dataset was used in the simulation phase, which can be downloaded from [26]. The data was captured from a three-axis accelerometer sensor which was sampled at 50 Hz [27]. Each data recorded in the dataset has a time window of 6s [27]. The ADL data are captured from real scenarios and fall activities are simulated [27]. The following fall activities are simulated: forward falls, backward falls, left and right lateral falls, syncope, sitting on an empty chair, preventing falling or reducing the impact of fall [27]. The ADL was carried out in real conditions, where the volunteers placed the phone in the trouser pocket for seven days to record their movements [27]. The fall experiment was selected from Nourey and Kangas [27]. Each of the fall activities was conducted three times, which makes a total of 24 simulations per participant [27]. Falls were performed on a soft mattress in a laboratory environment where the subjects placed the smartphone in both their left and right pockets [27]. The advantage of using this dataset compared to the other dataset is that the ADL is captured in a laboratory environment, and it represents subjects' real movements in their everyday life. The disadvantage of the dataset is that the type of ADL performed is not known. From the dataset, nine subjects were used in the simulation. For each record in the dataset, the time index of the maximum peak of the signal magnitude vector (SMV) is found. A fall acceleration signal comprises of peaks and valleys, and fall activities usually associated with large SMV peaks [28], [29]. The SMV equation is given below:

$$SMV = \sqrt{\ddot{x}^2 + \ddot{y}^2 + \ddot{z}^2} \qquad (1)$$

where $\ddot{x}, \ddot{y}, \ddot{z}$ are the acceleration values along the x, y and z axis of the accelerometer [30], [31]. From the time index of the SMV peak, where 2 seconds (50 Hz ×1 second = 50 samples) of data from the left and right the peak is extracted. A sliding window of the signal is captured for 101 values since the duration of the fall is about 2 seconds [32]. This reduces computational complexity when training and testing the models and machine learning algorithms [32]. From the time index of the SMV peak, where 2 seconds (50 Hz×1 second = 50 samples) of data from the left and right the peak is extracted. A sliding window of the signal is captured for 101 values since the duration of the fall is about 2 seconds [32]. The raw acceleration values, the x-axis, y-axis, and the z-axis from the accelerometer is extracted from the sliding, and the following vector is created,

$$Vector = [x_1, x_2, ..., x_{101}, y_1, y_2, ..., y_{101}, z_1, z_2, ..., z_{101}] \quad (2)$$

No feature extraction and feature selection methods were applied to the raw acceleration values. The raw acceleration values consist of the orientation of the smartphone and the three-dimensional information, which can be used to differentiate between ADLs and fall activities [27]. The advantage of this approach is that it saves processing time, and reduces computational complexity. The most popular supervised machine learning algorithm support vector machine (SVM) is used in Model 1, and angle-based outlier detection (ABOD) is used in the personalized model.

*1) Support Vector machine*: The SVM uses a kernel trick where it transforms the inputs which are features extracted into a higher dimensional space using a non-linear mapping in which an optimum hyperplane is found separating two classes from a given training dataset [33], [34], [35]. A hyperplane is used to separate the two classes by creating a decision boundary (maximum margin hyperplane) [11]. Optimization of separating hyperplane is done by maximizing the distance between the hyperplane and the nearest data points [3], [34]. The maximum margin hyperplane is learnt based on the support vectors, which the classifier used to classify the new feature vector [11], [33]. The Radial Basis Functions (RBF) kernel is the most popular kernel used in SVM [33]. The classifier makes use of hyperplane as a decision boundary to classify the binary data, which is given by [11], [33],

$$K(x_i, x) = \exp(-\gamma \| x - x' \|^2) \quad (3)$$

where $b$ is the bias term, K($x_i$; x) is the radial base function. Equation (4), gives the formula for the radial base function used.

$$K(x_i, x) = \exp(-\gamma \| x - x' \|^2) \quad (4)$$

*2) Angle-based outlier degree*: The increase in the dimensionality of feature space causes the relative contrast between distances to become smaller [36], [37]. The idea of the neighbourhood becomes useless if the dimensionality of feature space increases [36]. This is resolved by the use of angles rather than distances in high dimensional space since angles are more stable [36], [37]. The basic idea of ABOD is that outliers are located at the end of the data distribution and the normal points are in the centre of the data distribution [36]. The advantage of this method is that it does not need any parameter selection [36]. The distance is used only to normalize the results. [36]. To detect if the incoming data is an outlier, the variance of the angles between the different vectors in the dataset is calculated [36]. An outlier is detected if the angle is small [36]. Figure 4 shows the input data point (A), the angle between $\vec{AB}$ and $\vec{AC}$ for any two B, C data points from dataset (D). The ABOD factor is calculated as the variance angles between the differences vectors of the A to all pair of points in the dataset (D) weighted by the distance of the points. The reason for dividing by distance is for the low dimensional datasets where angles are less reliable [36], [37].

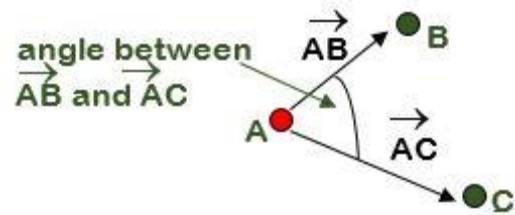

Fig. 4: Example of the ABOD model calculation of the angle between two points.

## 8 simulation results

To train and test Model 1 is shown in Figure 5 where SVM classifier is trained with all the data except the person who is going to be tested. The process is done 9 times since there are 9 people data in the dataset. The evaluation of personalized model is shown in Figure 6, where a 10-fold cross-validation is used, which divides the entire dataset into 10 parts, and takes one part for testing each time and the rest of the parts for training. This ensures that the entire dataset is tested. The cross-validation is applied only on person $p$ ADL records, where one-part is used for validation and the rest for training. From each person in the dataset except for $p$, 50 random ADL records are extracted which will create an initial training dataset of 400 ADL records. The reason for not using all the records from each person except for $p$, is to determine whether personalizing the system will improve the performance of the fall detection system.

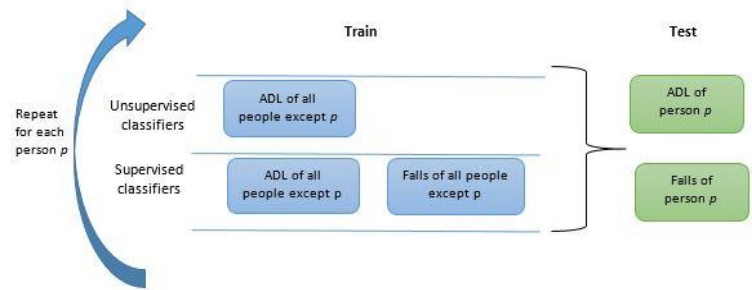

Fig. 5: The method used to train and evaluate Model 1.

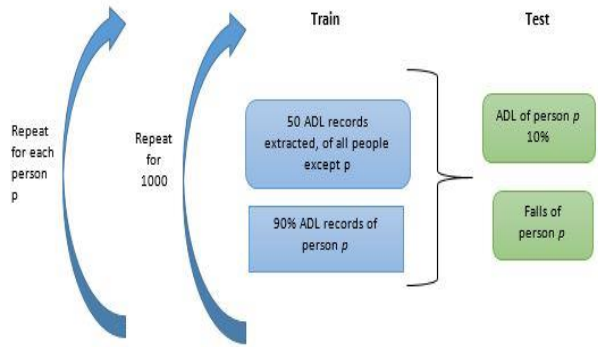

Fig. 6: The method used to train and evaluate the personalized model.

This is repeated 1000 times because random ADL records were extracted to create the initial training dataset.

To evaluate the performance geometric-mean will be used, given by

$$geometric\ mean = \sqrt{specificity \times sensitivity} \quad (5)$$

Specificity calculates how well a fall detection system can correctly detect ADL over the whole set of ADL instances [11], [24], [38], [39]. Sensitivity calculates how well a fall detection system can correctly detect falls over the whole set of fall instances [11], [38], [39].

Figure 7, presents the comparison between the common approach algorithm (Model 1) to the personalized fall detection algorithm based on the geometric mean. From the figure, personalized fall detection algorithm achieved best geometricmean.

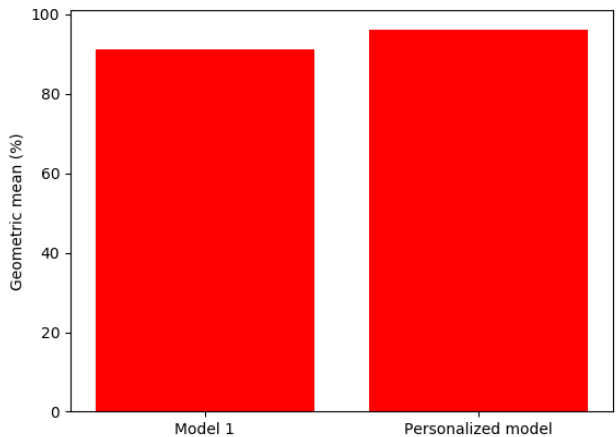

Fig. 7: Geometric-mean for Model 1 and personalized model using the tFall dataset.

## 9. Experimental

The aim of this experiment is to validate the system to determine if the most common ADLs can be detected. Since the tFall dataset used does not mention the type of ADL this will be implemented on the Android device. The phone captures the x, y, and the z-axis accelerometer data from the smartphone device. The data is sampled at 50 Hz, which is sufficient to detect if a fall has occurred [40]. Since an Android smartphone device is used, which is a battery-powered device, this requires low computational complexity algorithms [41].

To reduce the processing power and computation time, the following steps need to be taken before the classifier detects if an actual fall has occurred. The first step is to detect if a possible fall has occurred if the SMV is greater than 14.7. The second step is to detect if the angle of the smartphone is greater than 55 degrees; which indicates a lying state. The angle is calculated using only the x and y-axis of the accelerometer. If the second step is met, the raw acceleration data is sent to the ABOD classifier. The final algorithm implemented on the Android smartphone is shown in Figure 8.

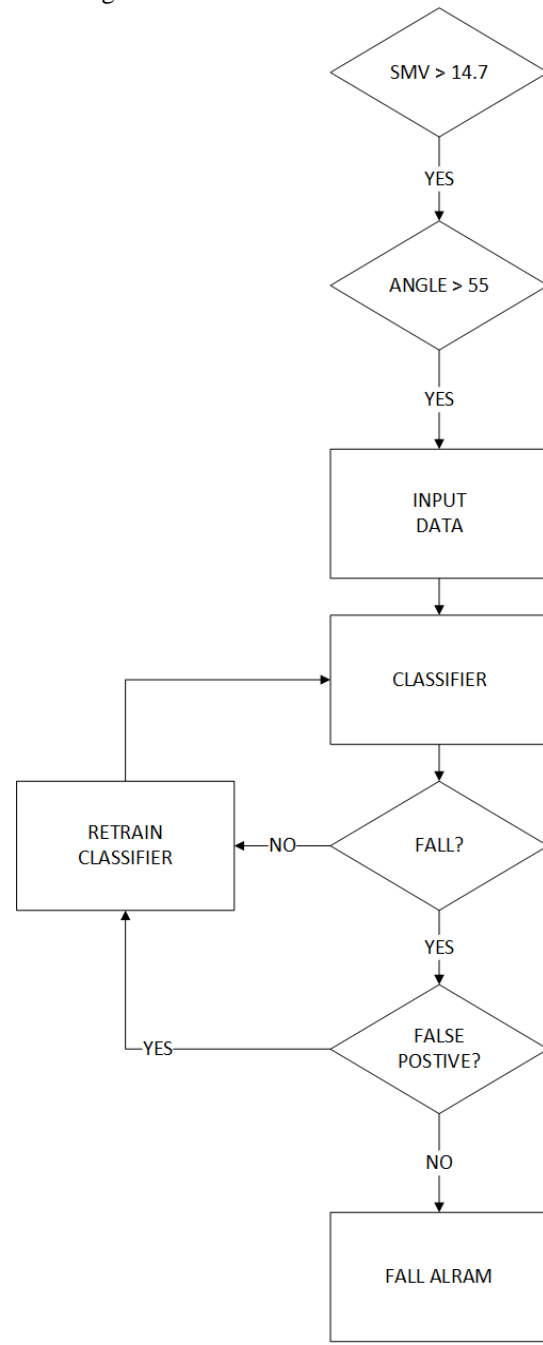

Fig. 8: Algorithm used on the android application for detecting falls.

TABLE I: ADL and fall activities movements list.

| ADL's experiments | | | Fall experiments | | |
|---|---|---|---|---|---|
| # | Label | Description | # | Label | Description |
| 1 | Walking fw | Walking forward | 41 | Front lying | From a standing position, going forward to the ground |
| 2 | Walking bw | Walking backward | 42 | Back lying | From a standing position, going backward to the ground |
| 3 | Jogging | Running | 43 | Rolling out bed | From a lying position, rolling out, ending on the ground |
| 4 | Stairs up | Going up the stairs (3 stairs) | 44 | Back sitting | From a standing position, to the ground while trying to sit on a chair |
| 5 | Stairs down | Going down the stairs (3 stairs) | 45 | Front knees lying | From standing position, to knees, to lying on the ground |
| 6 | Sit chair | Sitting on a chair | 46 | Right side | From standing position, going down on the ground, ending in right lateral position |
| 7 | Sit sofa | Sitting on a sofa | 47 | Left side | From standing position, going down on the ground, ending in left lateral position |
| 8 | Sit bed | Sitting on a bed | | | |
| 9 | Lying bed | From standing to lying position on a bed | | | |
| 10 | Pick object | Bending down to pick up an object on the floor | | | |
| 11 | Reach object | Standing on the tip of the toe to reach for an object in the cupboard | | | |
| 12 | Cough | Sneezing | | | |
| 13 | Jumping | Continuous jumping | | | |

Young and healthy adult subjects were used to test the fall detection system; since it is very difficult to test the falls on an elderly person. The following move-sets in Table I will be tested on four adults. The Android smartphone was carried by the four subjects for three days, placed in the trouser right pocket of the subject. The reason for this was to ensure that the fall detection system can be personalized to the subject's movements and record new activities. The subjects were requested to incorporate the ADL moveset from Table I, into their everyday lives. Each move-set will be performed three times, thus every subject will perform 39 ADL move-sets and 21 fall move-sets which brings the total to 60 move-sets for each subject. The subjects will be three men and one woman; ages of 24+/-2, body mass between 60 - 90kg, and the height between 1.5 - 1.81m. The user will perform these experiments on a 20cm soft mattress wearing a helmet, wrist and knee guards. To evaluate performance, the following metrics will be used; geometric-mean, sensitivity, and specificity. When the android application starts up, the following screen in Figure 9 is shown. The toggle button was on when the subjects were carrying the smartphone in the pocket for three days. In Figure 9, the fall and false alarm 7 button are disabled; it is enabled when the fall detection system detects a fall shown in Figure 10. In Figure 10, the subject can select the fall button when the fall has occurred or the false alarm button when no fall has occurred and that activity data is then sent to the classifier for retraining.

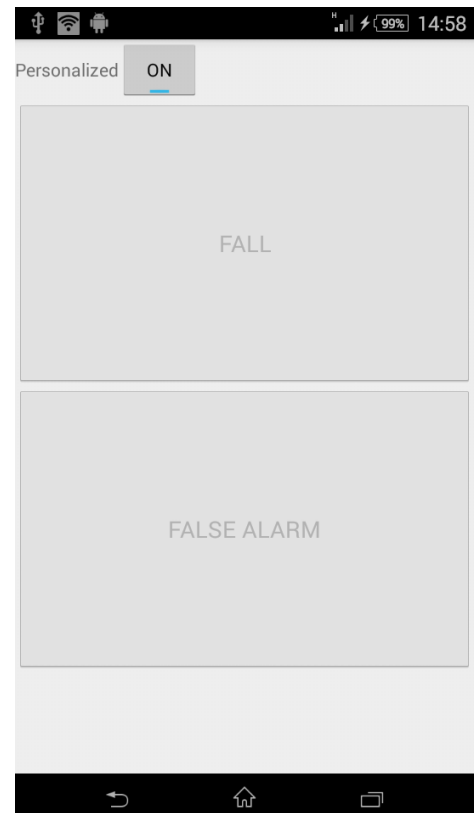

Fig. 9: Start-up screen when the Android application opens.

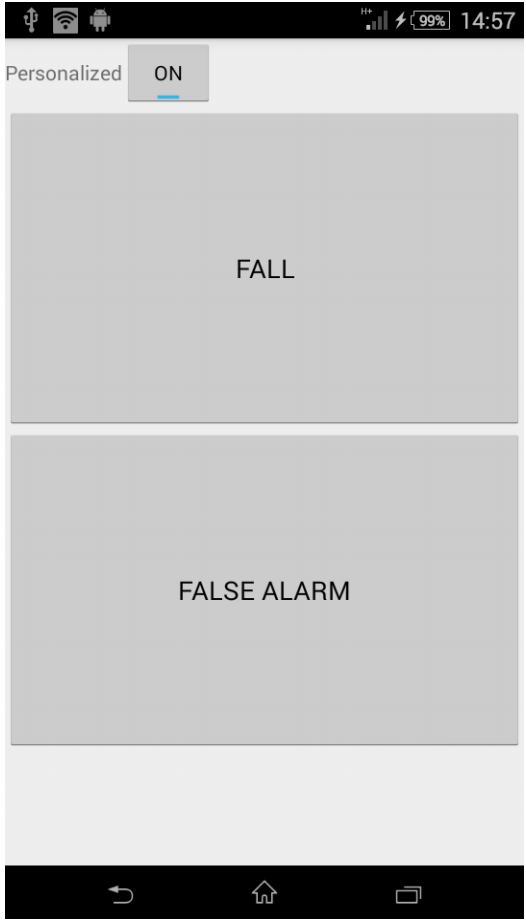

Fig. 10: When a fall has been detected by the Android application.

## 10 Experimental results

The results of the experiment are shown in Table II, where the values of SE, SP, their geometric-mean is recorded. About 80% of cells are higher than 90%. The mean values are above 90% (last row, bold face).

TABLE II: Values of SE, SP, and geometric mean for each subject is captured. The last row is the average value over subjects.

| Person | SP(%) | SE(%) | Geometric mean (%) |
|---|---|---|---|
| 1 | 92.31 | 85.71 | 88.95 |
| 2 | 89.74 | 90.48 | 90.11 |
| 3 | 94.87 | 90.48 | 92.66 |
| 4 | 92.31 | 95.24 | 93.76 |
| **Mean** | **92.31** | **90.48** | **91.37** |

## 11 Discussion

Based on the simulation results, personalization model outperforms the common fall detection model. This shows that future fall detection systems should only require one set of data, which is ADL. The reason for the high accuracy for SVM is that it has decision boundary to differentiate between ADL and fall activities. SVM achieves high accuracy in the experimental setup, but not in reality since it does not have the true fall event data. The reason for SVM not achieving higher accuracy compared to the personalized models is that the performance of the SVM classifier is highly dependent on the selected features. The purpose of the experimental test was to test how personalization would fare in everyday activities.

From the results, person 1 SE was below 89% due to hesitation when performing the fall activities; which resulted in the fall pattern being similar to an ADL pattern. Overall, the mean values are greater than 90% which is very good considering the phone was placed in the pocket, and more ADLs were added to the dataset compared to previous studies. The specificity and sensitivity are almost equal which shows that the fall detection system equally detects ADL and fall activities. The placement of the sensor may affect the accuracy of the system if not placed in the correct position or if the sensor is worn wrong [41]. The placement of the sensor around the waist produces low acceleration forces, hence low frequency which can result in low power consumption. There are not a lot of frequent movements around the waist region as it is the centre of gravity [41], [42], [43]. The reason for the low sensitivity may be that when test subjects perform the fall they hesitate to fall properly, due to danger. ADL accuracy is influenced by the dataset created; if the dataset does not contain that particular activity, the system generates a false positive. The accelerometer sensors can affect the performance of fall detection because different sensors have different noise levels. The range and resolution of the sensors in the smartphone can affect the performance of the fall detection system [40]. Phones were used in the experiment in this paper and the dataset was low cost where the accelerometer range is about ~2g; using a larger accelerometer range can reduce false alarms [27]. The reason being ~2g accelerometer cannot clearly differentiate acceleration peaks caused by the impact during a fall [40].

## 12 Advantages of personalized model

User-specific personalization can be provided using unsupervised machine learning algorithms, resulting in the following advantages: a) more activities can be included in the classifier, and b) the fall detection system can address the interindividual differences. If different body postures are not learnt, a high false rate could occur [19]. Only ADL data is required. Most fall detection systems require the device to be placed in a specific location on the person, with personalization the user can place the device on any recommended locations and gather the data. To get accurate fall data, a long-term experiment needs to be conducted in nursing homes using wearable sensors, ambient sensors, or camera-based methods [20]. This can be solved through personalization.

## 13 Conclusion

Personalized fall detection system is shown to provide added and more benefits compare to the current fall detection system. The personalized model can also be applied to anything where one class of data is hard to gather. The results show that adapting to the user needs, improve the overall accuracy of the system. Future work includes detection of the smartphone on the user so that the user can place the system anywhere on the body and make sure it detects. Even though the accuracy is not 100% the proof of concept of personalization can be used to achieve greater accuracy. The concept of personalization used in this paper can also be extended to other research in the medical field or where data is hard to come by for a particular class. More research into the feature extraction and feature selection module should be investigated. For the feature selection module, more research into selecting features based on one class data.